%% file: main.tex
\documentclass{article}

\usepackage{amsmath}

\usepackage{microtype}
\usepackage{graphicx}
\usepackage{subfigure}
\usepackage{booktabs} %

\usepackage{hyperref}

\usepackage[accepted]{icml2020}

\usepackage{xfrac}
\usepackage{todonotes}
\usepackage[nameinlink]{cleveref}  %

\newcommand{\DD}{{\cal D}}
\newcommand{\LL}{{\cal L}}

\newcommand{\HH}{{\cal H}}
\newcommand{\MDL}{\text{DL}}

\definecolor{CommentCyan}{rgb}{0,.8,.8}
\definecolor{CommentCyan2}{rgb}{0,.6,.6}
\definecolor{CommentRed}{rgb}{0.9,0,0}
\definecolor{CommentGreen}{rgb}{0,0.7,0}
\definecolor{CommentBlue}{rgb}{0,0,0.7}
\definecolor{CommentDarkBlue}{rgb}{0,0,0.4}
\definecolor{CommentOrange}{rgb}{0.8,0.6,0.0}
\definecolor{CommentOrange2}{rgb}{0.6,0.4,0.0}
\definecolor{CommentViolet}{rgb}{0.3,0.,0.5}

\newif\ifshowcomments

\showcommentsfalse

\newcommand{\jorg}[1] {{\ifshowcomments\color{CommentCyan} {\footnotesize{{[Jorg: \textbf{#1}]}}}\fi}}
\newcommand{\simon}[1] {{\ifshowcomments\color{CommentRed} {\footnotesize{{[Simon: \textbf{#1}]}}}\fi}}
\newcommand{\fra}[1] {{\ifshowcomments\color{CommentGreen} {\footnotesize{{[Fra: \textbf{#1}]}}}\fi}}

\icmltitlerunning{Small Data, Big Decisions: Model Selection in the Small-Data Regime}

\begin{document}

\twocolumn[
\icmltitle{Small Data, Big Decisions: Model Selection in the Small-Data Regime}

\icmlsetsymbol{equal}{*}

\begin{icmlauthorlist}
\icmlauthor{J{\"o}rg Bornschein}{dm}
\icmlauthor{Francesco Visin}{dm}
\icmlauthor{Simon Osindero}{dm}
\end{icmlauthorlist}

\icmlaffiliation{dm}{DeepMind, London, United Kingdom}

\icmlcorrespondingauthor{J{\"o}rg Bornschein}{bornschein@google.com}

\icmlkeywords{Generalization, small-data}

\vskip 0.3in
]

\printAffiliationsAndNotice{}  %

\begin{abstract}

Highly overparametrized neural networks can display curiously strong generalization performance -- 
a phenomenon that has recently garnered a wealth of theoretical and empirical research in order 
to better understand it.
In contrast to most previous work, which typically considers the performance as a function of the model size, in this paper we empirically study the generalization performance as the size of the training set varies over multiple orders of magnitude. 
These systematic experiments lead to some interesting and potentially very useful 
observations; perhaps most notably that training on smaller subsets of the
data can lead to more reliable model selection decisions whilst simultaneously enjoying smaller computational overheads.
Our experiments furthermore allow us to estimate Minimum Description Lengths for common datasets given modern neural network architectures, thereby paving the way for principled model selection taking into account Occams-razor.

\end{abstract}

\input{01-intro.tex}

\input{02-methods.tex}

\input{03-experiments.tex}

\input{04-mdl.tex}

\input{05-conclusions.tex}

\bibliography{main}
\bibliographystyle{icml2020}

\input{supplement.tex}

\end{document}

%% file: 01-intro.tex
\section{Introduction}

\begin{figure*}[t]
    \centering
    \includegraphics[width=\textwidth]{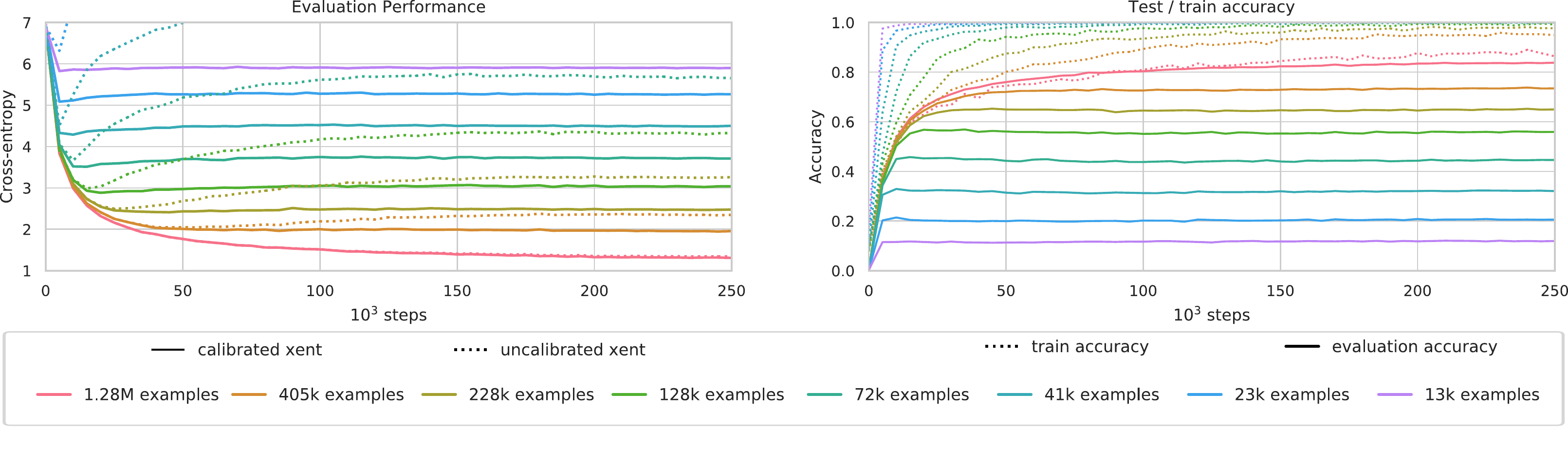}
    \vspace{-2em}
    \caption{Learning curves for ResNet-101 on subsets of 
    the ImageNet dataset using a RMSProp with a cosine learning-rate schedule.
    Left: the uncalibrated generalization cross-entropy shows a strong overfitting signature (dashed line), the calibrated cross-entropy does not.
    }
    \label{fig:learningcurves}
\end{figure*}

\begin{figure}
    \centering
    \includegraphics[width=\columnwidth]{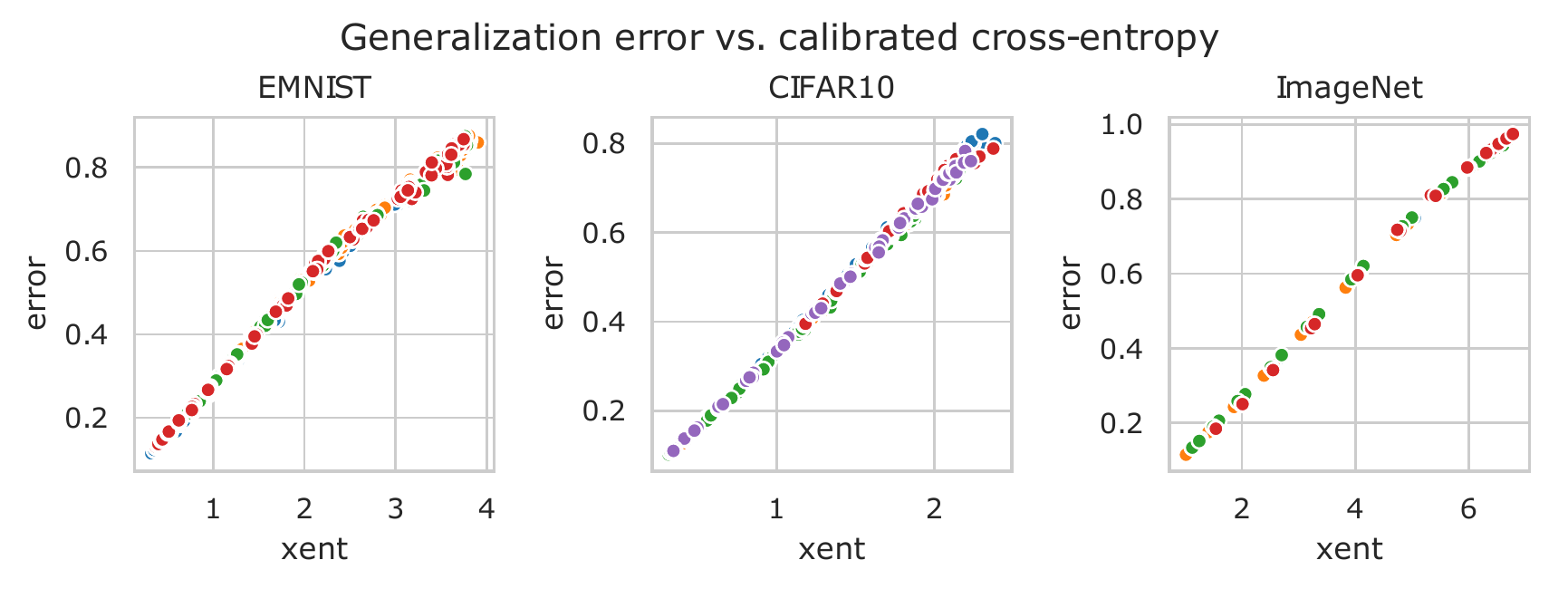}
    \vspace{-2em}
    \caption{Post-convergence generalization error vs. generalization cross-entropy 
    without early-stopping for a range of model architectures  and 
    training set sizes. 
    The colours represent the different model architectures. 
    We observe that the calibrated cross entropy is strongly 
    correlated with the generalization error rate.}
    \vspace{-0.7em}
    \label{fig:err-vs-xent}
\end{figure}

According to classical statistical learning theory, achieving an optimal generalisation loss requires selecting a model capacity that strikes the best balance between underfitting and overfitting, i.e., between not having enough capacity to model the training data accurately and having too much, and thus prone to adapt too closely to the training data at the expense of generalisation. Under this theory, the final generalisation loss plotted against model capacity should behave as a U-shaped curve -- initially decreasing as the capacity increases (underfitting) to reach a minimum (optimal model size) and then finally increase again (overfitting).

Contrary to these results, it has long been observed that neural networks
show a curiously good generalization performance (in terms of error) when applied to classification problems, 
even though the generalization cross-entropy exhibits all the characteristics of overfitting.
There has recently been renewed interest in studying this phenomenon, both theoretically and empirically
\citep{advani2017high, spigler2018jamming}.
\citet{belkin2019reconciling}  %
for example 
argue that beside the classical underfitting and overfitting regimes, a third one for massively overparameterised models exists.
The transition into this regime is called the {\em interpolation threshold} and is characterized by a peak in 
generalization error, followed by a phase of further decrease. The peculiar shape 
of the generalization-error over model-size curve lends the term {\em double-descent} to this phenomenon.
Work by \citet{nakkiran2019deep} has sharpened this picture and shown that this behaviour can be observed when training modern neural network architectures on established, challenging datasets.
Most of the work studying generalization of neural networks focuses on the relationship between generalization performance and model size. 
Instead we present an empirical study that investigates the generalization performance as a function of the training set size.

In the rest of the Introduction we outline the main contributions of our work.

\subsection{Performance analysis} 

One of our key contributions is to gather performance curves as a function of the training set size for a wide range of architectures on ImageNet, CIFAR10, MNIST and EMNIST. We also perform an extensive sweep over a wide range of architectures, model sizes, optimizers and, crucially, we vary the size of the training sets over multiple orders of magnitude, starting from the full dataset down to only few examples per class. Our study covers even extreme cases -- for example the training of oversized ResNet architectures with 10 or less examples per class.
This is, to the best of our knowledge, the most extensive empirical analysis conducted on 
generalization for massively overparameterized 
models, and strengthens the emerging understanding of training regimes for modern deep learning.

\subsection{A ranking-hypothesis}

One salient observation we have made has not been described in the literature: overparameterized model architectures seem to maintain their relative ranking in terms of generalization performance, when trained on arbitrarily small subsets of the training set. 

This observation prompts us to hypothesize that:  when (i) two sufficiently
large neural network architectures A and B are trained with a well 
tuned optimizer on datasets of size N; and (ii) we observe that, in expectation,  
A performs better than B; then (iii) A will also perform better than B in expectation for all differently sized datasets drawn from the same underlying distribution, as long as we remain well beyond the interpolation threshold. 

Unfortunately this is only an hypothesis, but if this conjecture is true it would have profound practical 
implications. Namely, it would mean that, for sufficiently large models, it would be possible to perform
model selection or architecture-search using small subsets of the data, and expect that
the decision regarding which model to prefer remains valid when applied to much larger datasets.
Indeed, our experiments show that training on small or medium sized subsets of the training data leads in many cases not only to faster convergence, but also 
to a more robust signal for model selection than training 
on big datasets, and is therefore often preferable.

\subsection{Calibration \& minimum description length}
Independently of the model-selection hypothesis, we also show that it is possible to 
avoid some negative effects of overfitting by simply choosing an optimal 
softmax-temperature on a small held-out dataset; i.e., by calibrating 
the neural network \citep{guo2017calibration}.
After calibration, the generalization cross-entropy becomes a stable and 
well-behaved quantity even when model sizes and training set sizes vary 
considerably. Being able to compute well-behaved generalization cross-entropy
on small training sets allow us to compute reliable Minimum Description 
Length estimates, a quantity that is of interest for principled model 
selection taking into account Occam's-razor. We will discuss this more in depth in \Cref{mdl}.

\section{Related Work}

The literature on generalization performance for learned predictors is vast 
and spans seminal work on the classical bias-variance-tradeoff
\citep{geman1992neural, domingos2000unified} all the 
way to theoretical and empirical work investigating the still poorly 
understood, but often strong generalization performance of 
neural networks.
The latter has been approached from a number of different directions: For example
pointing out that neural networks seem to perform implicit capacity 
control \citep{zhang2016understanding}, investigating the learning dynamics 
and properties of the loss landscape \cite{spigler2018jamming} and 
interpreting stochastic gradient descent as an approximation of probabilistic inference 
\citep{mandt2017stochastic}.
Finally, two lines of work recently contributed to the understanding of generalization, the first studying infinitely wide neural networks
\citep{jacot2018neuraltangent, allen2019learning} and
the second focusing on the double-decent phenomenon 
\citep{belkin2019reconciling, nakkiran2019deep}.

The vast majority of these previous studies focus on the model-size dependency 
of the generalization performance. 
Notable exceptions are the work by \citet{nakkiran2019deep}, who investigates 
the neighbourhood of the interpolation threshold, 
and the work by \citet{hestness2017deep}, who 
first evaluated the generalization error-rate for language models 
and ResNets as a function of the training set size. 
They point out that the generalization error-rate can be well predicted  by 
assuming a power law and interpolating from smaller training sets to bigger ones. 
In a similar fashion, our work shows that the relative model ranking trained on small datasets is maintained when trained on bigger ones, a result of practical importance for model selection and architecture search.

%% file: 02-methods.tex
\begin{figure*}[t]
    \centering
    \includegraphics[width=\textwidth]{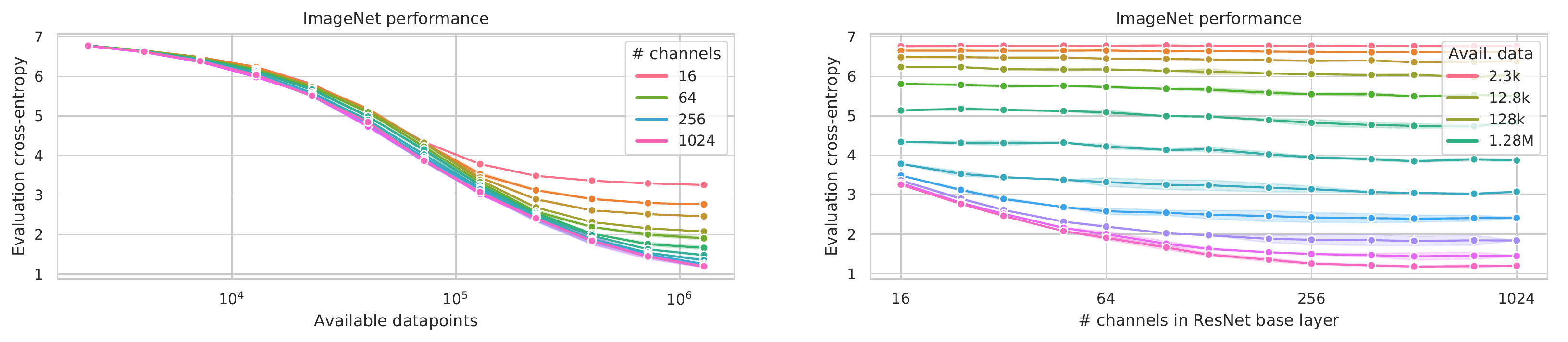}
    \vspace{-2em}
    \caption{Cross-entropy performance profiles for the ResNet-101 architecture on ImageNet when trained with RMSProp. 
    We use 90\% of the available data for training, 10\% for calibration and report the generalization performance on 
    the unseen validation set. Note that even when training with as little as $\approx 2.3$ images per class, there is no 
    harm in using a ResNet model with 4 $\times$ more channels (16$\times$ more parameters) than the standard 
    ResNet.
    }
    \label{fig:channels-imagenet}
\end{figure*}

\begin{figure}
    \centering
    \includegraphics[width=\columnwidth]{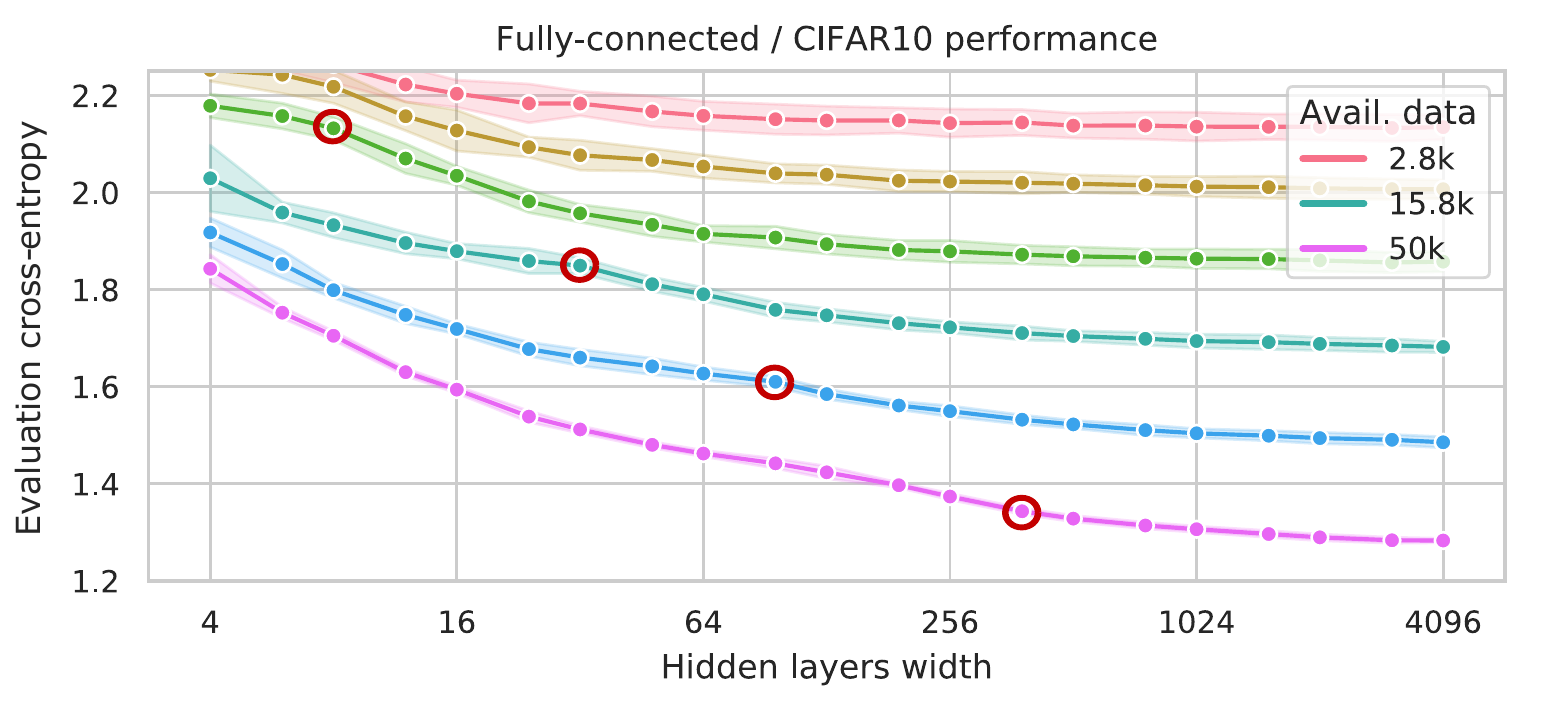}  %
    \vspace{-2em}
    \caption{Performance profiles for a fully connected MLP with 3 hidden layers on CIFAR10 as a function of the hidden 
        layer size. Red points mark the smallest models that approach a close to zero training error-rate.
    }
    \label{fig:channels-cifar10-mlp}
\end{figure}

\begin{figure*}[t]
    \centering
    \includegraphics[width=\textwidth]{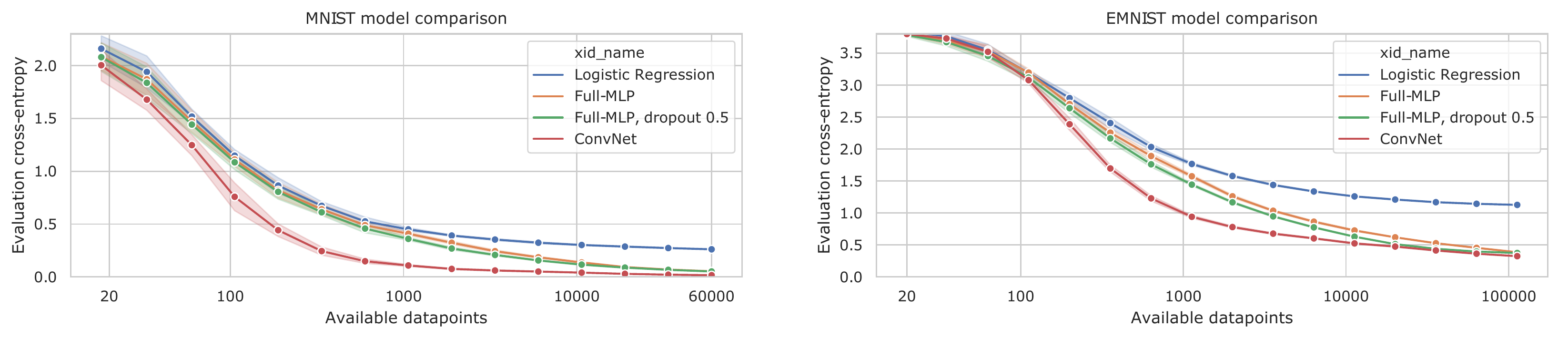}
    \\
    \includegraphics[width=\textwidth]{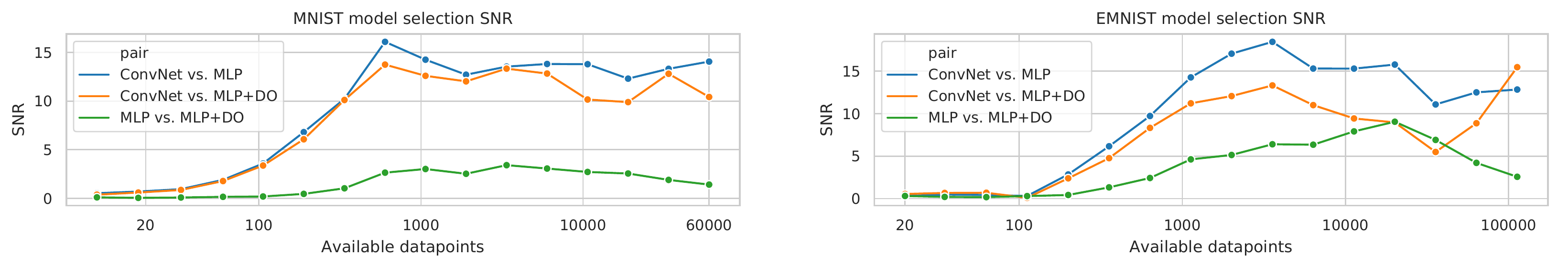}
    \vspace{-2em}
    \caption{
    {\bf Top row}: Generalization performance for various model architectures trained with Adam 
      on MNIST and EMNIST as a function of training set size. Uncertainty bands represent standard-deviation 
      after training with 30 different seeds.
    {\bf Bottom row}: SNR for the performance difference between models; estimated using 1000 bootstrap samples.
    }
    \label{fig:mnist-compare}
\end{figure*}

\begin{figure}[t]
    \centering
    \includegraphics[width=\columnwidth]{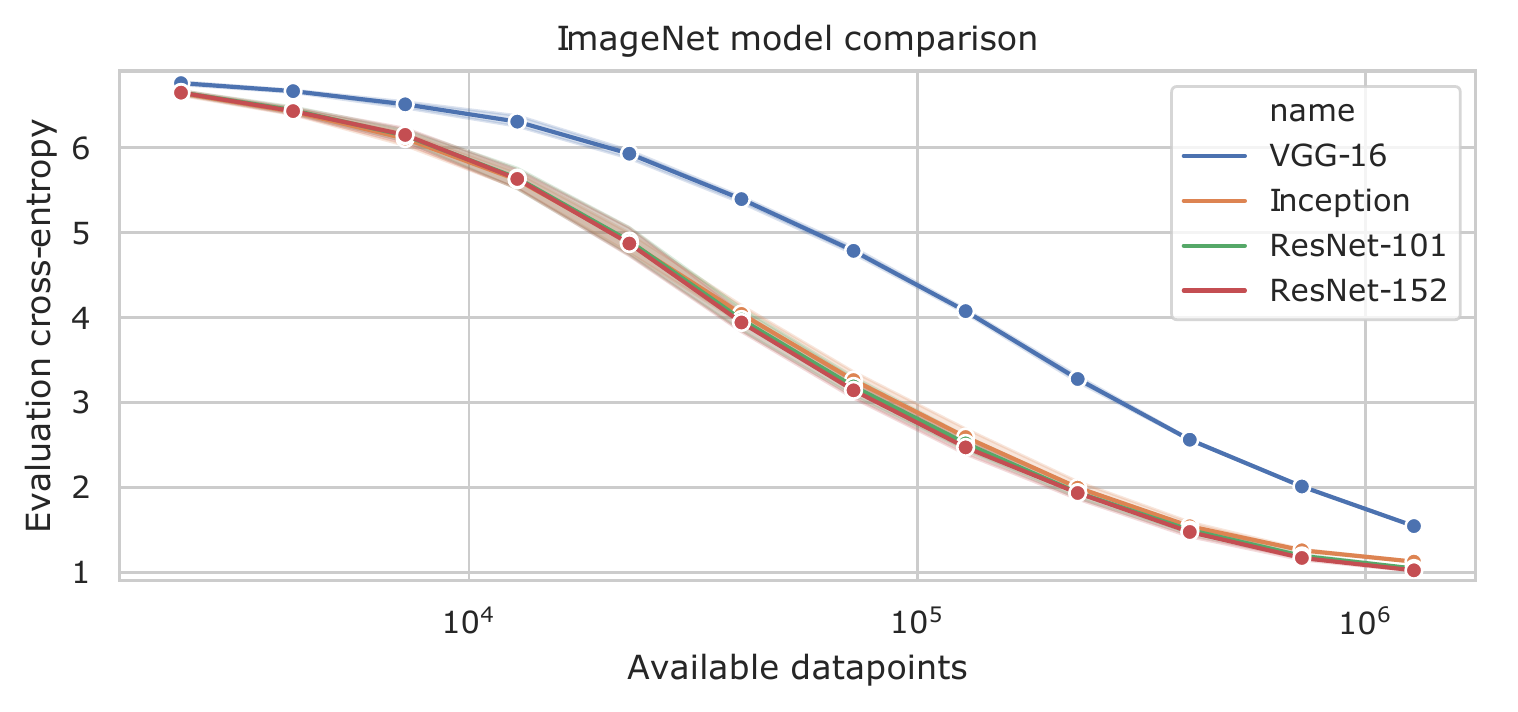}
    \\
    \vspace{1em}
    \includegraphics[width=\columnwidth]{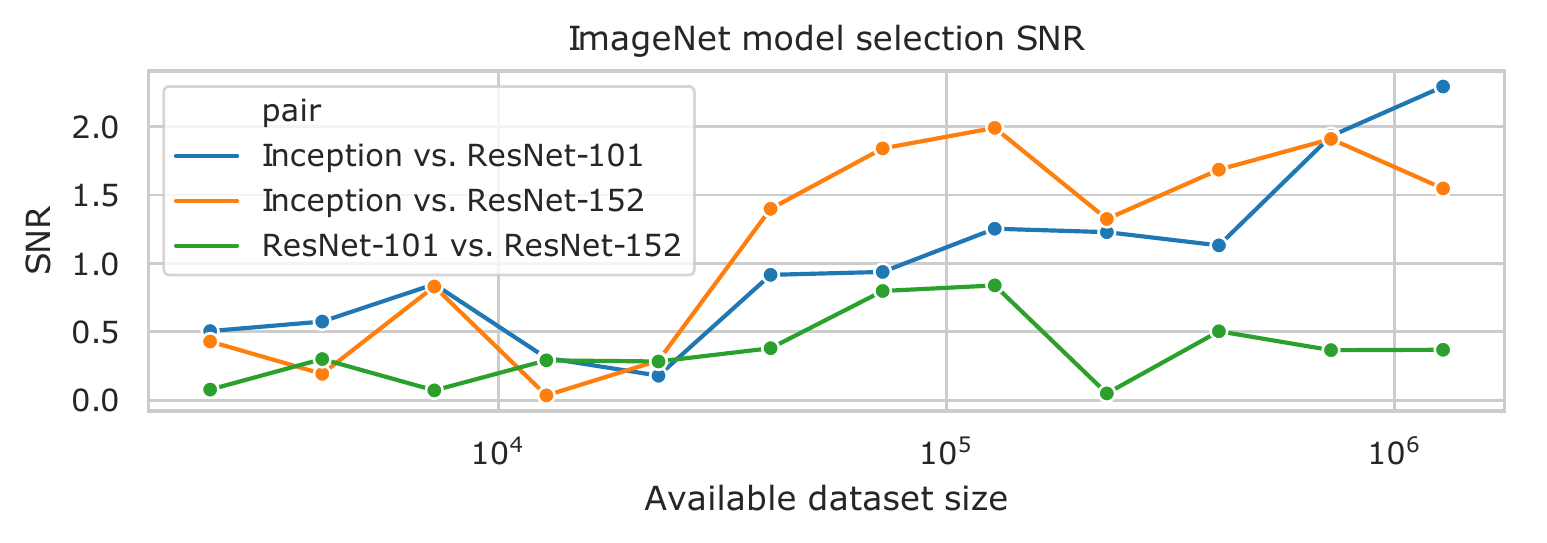}
    \vspace{-2em}
    \caption{
    {\bf Top}: Generalization performance for models trained on ImageNet; 
        Uncertainty bands from training 5 different seeds.
    {\bf Bottom}: SNR for the performance difference between models; estimated using 1000 bootstrap samples.
    }
    \label{fig:imagenet-compare}
\end{figure}

\begin{figure*}
    \centering
    \includegraphics[width=\columnwidth]{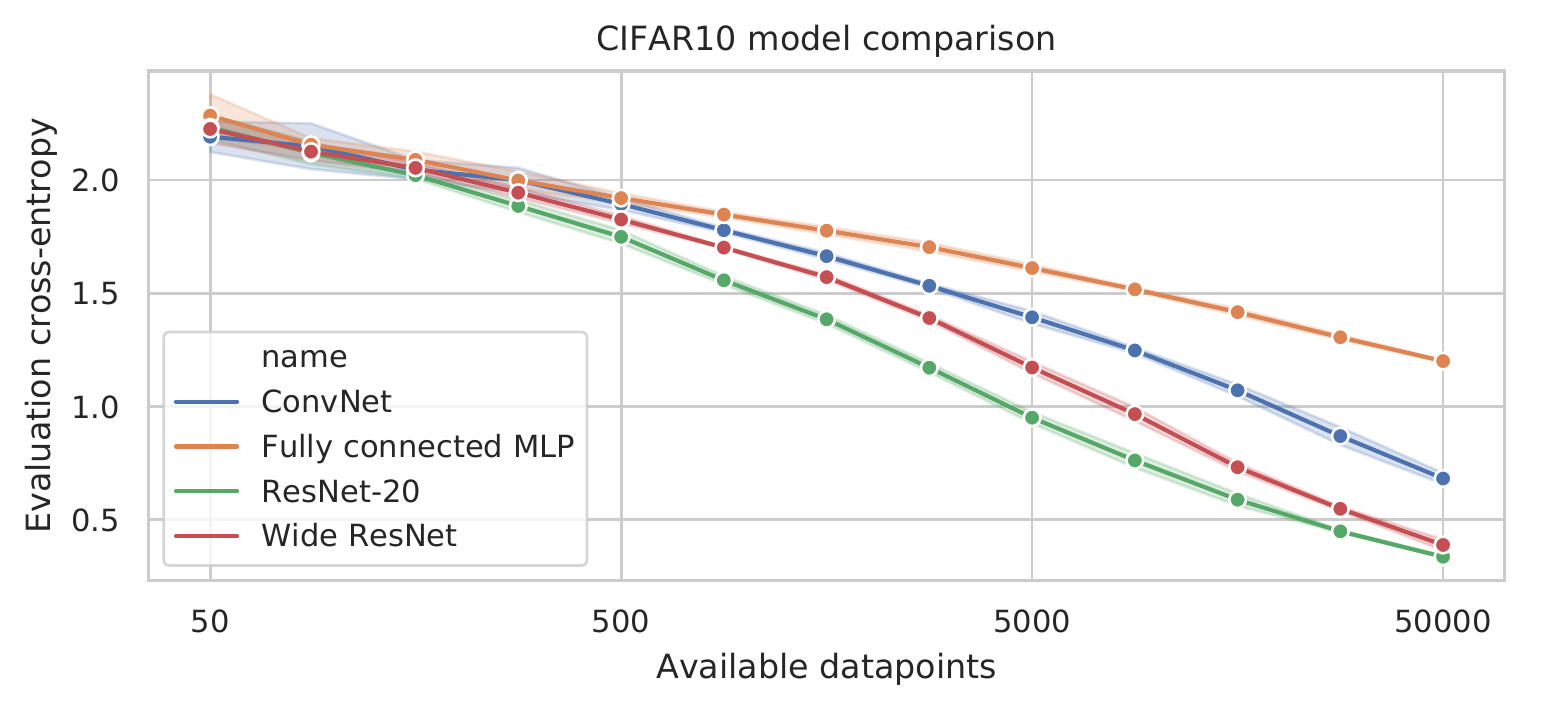}
    \includegraphics[width=\columnwidth]{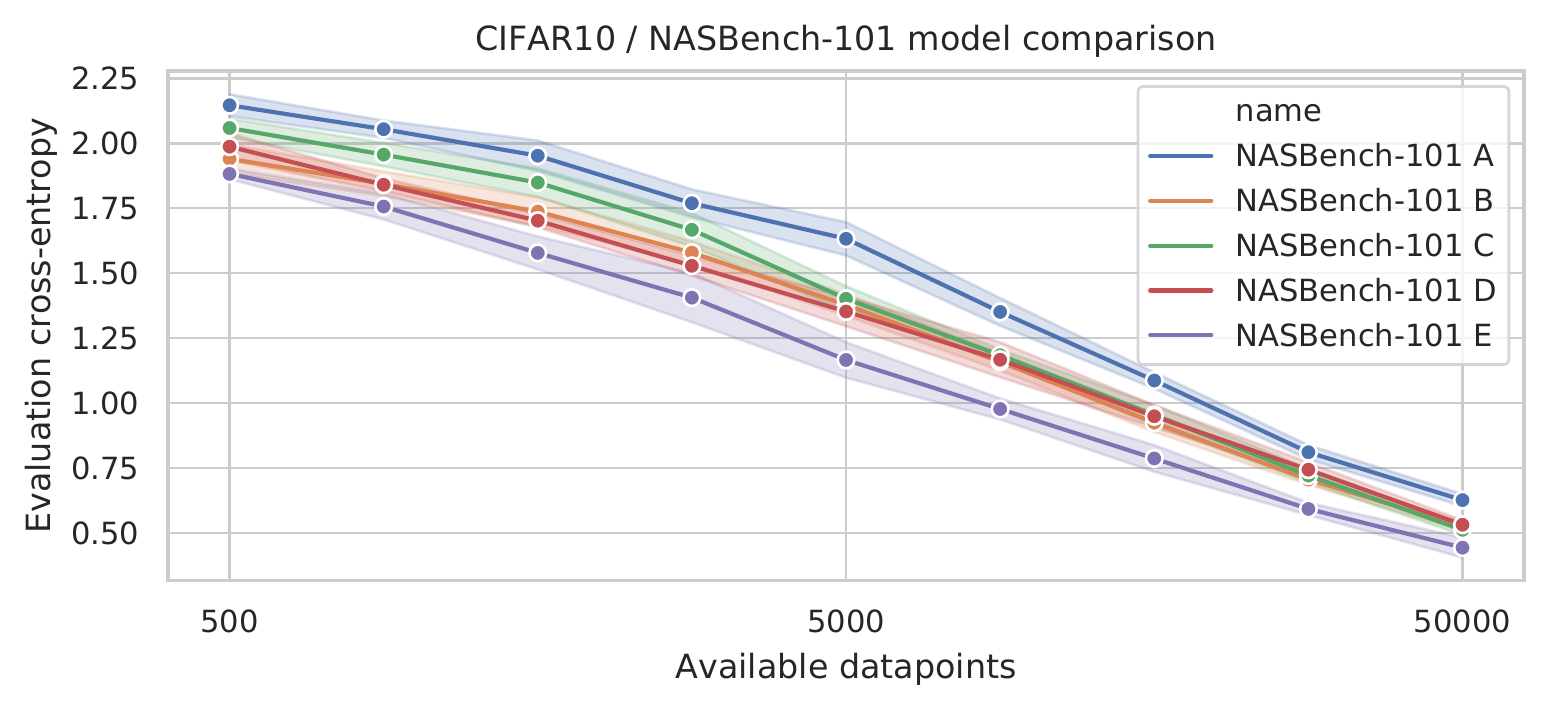}    
    \caption{
    {\bf Left}: Selected models trained on CIFAR10; uncertainty bands from 30 different seeds. The ConvNet has
        4 hidden layers with 3x3 kernels and 256 channels; every second layer uses stride 2 width a single 2048 unit 
        wide fully connected hidden layer on top.
    {\bf Right}: Five models equidistantly selected from the NASBench-101 architecture search space; 
    50\% of the available datapoints were used for training, 50\% for calibration. 
    }
    \label{fig:cifar10-compare}    
\end{figure*}

\section{Methods}

\subsection{Temperature calibration}
\label{calibration}

Overparameterized models are typically trained by minimizing
either a regression loss or a categorical cross-entropy loss,
but in contrast are then evaluated on their generalization performance measured on the error-rate.
This is because the generalization cross-entropy can overfit 
severely for models that are sufficiently big.

We show here that it is possible to avoid the negative effects of 
overfitting when using the categorical cross-entropy by simply choosing an optimal 
softmax-temperature on a small heldout dataset; i.e., by calibrating the neural 
network in the way proposed by \citet{guo2017calibration}.
In practice we implement temperature calibration by performing gradient descent 
on the calibrated cross-entropy loss w.r.t. a single scalar temperature parameter. 
We interleave the regular model training steps with steps of gradient descent on the calibration loss on some held-out data, 
that we refer to as the {\em calibration dataset}.
This allows us to track the generalization cross-entropy online during 
learning\fra{Can you specify why this is important?}.

This calibration procedure can prevent the generalization cross-entropy from overfitting, even when the model size is order of magnitudes larger than the training set size, and without relying on early-stopping (see \Cref{fig:learningcurves}
and \Cref{fig:err-vs-xent}).
Note that, just as with the post-convergence calibration proposed by
\citet{guo2017calibration}, the calibration procedure does not modify the other parameters, 
directly or indirectly.

Being able to compute well-behaved generalization cross-entropies is desirable because it
is the loss we optimize for and because it allows us to compute MDL estimates, 
as explained in \Cref{mdl}. That being said, we could express all other results in this 
paper in terms of error-rates instead of cross-entropies, and the observations and 
conclusions would still hold (see Supplementary material).

\subsection{Datasets and models}

Throughout this study we present a large number of experiments on 
subsets of different sizes of popular benchmarking datasets. 
We call the total set of datapoints available for a particular 
training run the {\em available dataset}, which is split into a {\em training set} and a
{\em calibration set}. The former is used to optimize the connection weights, the 
batch-norm parameters and all other parameters that are considered part of the neural networks, while the latter to optimise the calibration temperature, as explained in Section \ref{calibration}, as well as to determine the optimal learning-rate and, potentially, to perform early-stopping.
If not mentioned otherwise, we will use a 90\%/10\% training/calibration split of the \emph{available dataset}.
To assess a model's ability to generalise beyond the available-dataset, we then successively evaluate
 them on a separate held-out or {\em evaluation dataset}. 
We experimented with balanced subset sampling, i.e. ensuring that all subsets always contain
an equal number of examples per class. But we did not observe any reliable improvements from doing 
so and therefore reverted to a simple i.i.d sampling strategy. 
We always pay particular attention to not to inadvertently leak data, i.e. use datapoints that have 
not been properly accounted for to select the hyperparameters.

We conduct experiments on the following datasets:\\
{\bf MNIST} consists of 60k training and 10k test examples from 10 classes
\citep{lecun1998mnist}. 
We train MLPs of various depth and width, with and without dropout, as well as standard ConvNets 
on this dataset. Unless otherwise noted, we use ReLU as the nonlinearity. \\
{\bf EMNIST} provides 112,800 training and 18,800 test datapoints from 47 classes in its balanced subset
\citep{cohen2017emnist}. 
We train the same family of model architectures we also train on MNIST. \\
{\bf CIFAR10} consists of 50k training and 10k test examples from 10 classes
\citep{krizhevsky2009learning}. 
We train a wide range of models on this dataset, including simple  architectures 
like MLPs and ConvNets; architectures that have been carefully optimized for image 
classification: ResNet-20 \citep{he2016deep} and Wide ResNets \citep{zagoruyko2016wide} 
as well as a selection of architectures from the NASBench-101 \citep{ying2019bench} search space.
The latter were chosen by disregarding the worst 10\% and then picking 5 equally-spaced
from the remaining ones. The Supplementary material contains the description
of these architectures. The rationale is that we want to evaluate a range 
on non-optimal architectures and confirm their relative ranking is preserved when using smaller
datasets too. \\
{\bf ImageNet} contains 1.28M training and 50k validation examples from 1000 classes
\citep{ILSVRC15}. 
We train a selection of widely known standard models like VGG-16 \citep{simonyan2014very}, ResNets \citep{he2016deep}
and Inception \citep{szegedy2016rethinking}. Additionally we consider S3TA \citep{zoran2019towards}, 
a sequential and attention based model that takes multiple glimpses at various spacial locations in the
image before emitting a prediction. It is composed of a ResNet-style feature
extractor with a reduced number of strides, to keep the spatial resolution higher, 
and an LSTM equipped with a spatial attention mechanism to ingest features.
The rationale for including it in our exploration is that it supposedly has a 
different inductive bias than pure conventional models, which could increase the 
chance of falsifying our consistent ranking hypothesis. 

\subsection{Training}

To ensure our observations are not specific to a particular optimization 
method, we run experiments with different variants of gradient-descent.
For each experiment we sweep over a fixed set of possible learning rates 
and pick the best one according to the calibration loss, independently for
each model architecture and training set size.

Throughout this study we use the following optimizers: \\
{\bf Adam} \citep{kingma2014adam} with fixed learning rates 
$\{10^{-4}, 3 \cdot 10^{-4}, 10^{-3}\}$ and 50 epochs.\\
{\bf Momentum SGD} with initial learning rates $\{10^{-4}, 3 \cdot 10^{-4}, \cdots, 10^{-1}\}$ 
cosine-decaying over 50 epochs down to 0 (0.9 momentum and $\epsilon = 10^{-4}$). \\
{\bf RMSProp + cosine schedule} \citep{tieleman2012lecture} with initial learning rates 
of $\{0.03, 0.1, 0.3\}$ and cosine-decaying to 0 over 50 epochs. 
We choose the same hyperparameters used by \citep{ying2019bench} for 
their NASBench-101 experiments, with the exception of the number of epochs, which we reduced from 
108 to 50 (momentum=0.9, $\epsilon=1$).

For all experiments we use a batch size of 256 examples. 
The term epoch always refers to the number of gradient steps required
to go through the full-sized dataset once; i.e., on ImageNet an epoch is always $1.28M / 256 = 5000$ gradient 
steps, regardless of the size of the actual training set used.

We evaluated many more combinations than those presented in this paper, a selection of which is contained in the Supplementary material. Throughout all these control runs our qualitative results were confirmed, which suggests that these results do not just emerge from the interaction of specific 
models and optimizers; or from a particularly sensible hyperparameter choice.

%% file: 03-experiments.tex
\section{Experiments}

\subsection{Properties of calibration} 

\fra{fig 1 is too far away!}
\jorg{Think so too; on the other hand having a nice figure instead of a wall of text on page 2 leaves IMHO a good first impression; not sure how to prioritize}
\Cref{fig:learningcurves} visualizes typical learning curves when training a 
ResNet-101 model with cosine-decayed RMSProp on subsets of ImageNet. 
Depending on the training set size the model can memorise the training set and reach 
a zero training error-rate within a few thousand gradient steps. 
The uncalibrated cross-entropy on held-out data shows severe symptoms of overfitting in 
these cases but, as reported before, the error computed on the same set tends to instead flatline 
and not to degrade significantly. %

This suggests that the models are not unfavourably adjusting their decision boundaries as 
training progresses into the overfitting regime, but are first and foremostly becoming unduly 
confident in their predictions. When using calibration we rectify this over-confidence and  
obtain stable cross-entropies that show the same kind of robust behaviour as the evaluation error-rate.

Early-stopping can still have a positive effect though: We regularly 
observe a small degradation in generalization performance just around the point where the
rapid decrease in training error tapers off into a slow decrease towards zero. 
We can usually minimize this effect by making the models bigger and 
tuning the learning rate more accurately to reach optimal performance.
These observations are not specific to this experiment, but generalize 
to all optimizers and models considered in the paper. 

\Cref{fig:err-vs-xent} depicts the very strong correlation between the generalization error  and calibrated crossentropy for 
EMNIST models trained with Momentum-SGD, CIFAR10 models with Adam 
and ImageNet models with annealed RMSProp.

\subsection{Interaction of model size and training set size}

We designed a set  of experiments to directy connect to previous  
work and to confirm that the recent insight regarding model size and 
generalization performance holds in the small data regime.
Just like previous work we vary the model size by proportionally scaling 
the number of channels in all the convolutional layers or, for MLPs, the width of all the fully 
connected layers throughout the models. In contrast to previous work, here we additionally sweep over the size of available 
data used for training. 
As expected, it is evident from \Cref{fig:channels-imagenet}b) and \Cref{fig:channels-cifar10-mlp}
that scaling up models does generally improve the generalization performance, but comes with a price in terms of memory and computational effort, and is subject to diminishing marginal improvements. Selecting a model-size for a problem 
therefore requires finding the best trade-off between computational resources and the missed 
generalization performance one is willing to accept.
\fra{I think this is pretty obvious, I suggest to remove it.}
\jorg{No strong opinion; kind of like the clarification; but this is probably the first thing to throw out when we are out of space}
An interesting observation from these 
experiments is that the model size to achieve close-to-optimal performance 
seems to be independent of the size of the training set. 
For ResNet-101 on ImageNet for example (\Cref{fig:channels-imagenet} b),
it takes~$\approx$~384~channels in the first ResNet block to achieve close to optimal 
performance when training on either the 10k or the 1.2M samples training sets. Similarly on CIFAR10
(\Cref{fig:channels-cifar10-mlp}), between 1024 and 2048 units in the hidden layers achieves close-to optimal performance on all training sizes.
The Supplementary material contains additional plots suggesting the same constant relationship over a number of experiments.

According to the double-decent perspective \citep{belkin2019reconciling} 
we would expect the generalization error 
(and potentially the generalization cross-entropy) over model-size to spike around the point 
where a model barely reaches approximate zero training error.
We never quite see a spike, but we do observe recognizable artifacts in 
the curves; most pronounced on CIFAR10 when using a fully connected MLP 
(see \Cref{fig:channels-cifar10-mlp}).
For all other models and datasets this effect was much less pronounced and
usually not recognizable.

\subsection{Consistent Model Ranking}

Having established that we get reasonable and consistent results when
training big neural networks on tiny datasets down to only few 
examples per class, and that it is neither necessary nor desirable to
downscale the model size when the training set is small, 
we can now focus on comparing different model architectures on a given dataset.
For all these experiments we choose sufficiently big models: models that  
are beyond their interpolation threshold for the full datasets. 
More precisely, we consider a model ``big enough'' if doubling its size (in terms of number of 
parameters) does not result in a noticeable generalization improvement.

\Cref{fig:mnist-compare} (top-row) shows the results 
for MNIST and EMNIST. For both datasets we compare fully connected MLPs with 3 hidden 
layers, 2048 units each, either with or without dropout and a simple convolutional
network with 4 layers, 5x5 spatial kernel, stride 1 and 256 channels.
For comparison we also added logistic regression, even though we do not consider it 
part of our hypothesis because it can not be scaled to be a universal approximater to approach the 
irreducible Bayes-error.%

Overall, we tried many different variations and architectures: Replacing ReLUs with tanh non-linearities, adding batch-norm or layer-norm, changing the number of hidden layers 
or convolution parameters. 
Many of these changes have an effect on the performance over 
dataset-size curves, but we could not find a pair of architectures with performance
curves that crossed outside of their uncertainty bands, i.e., whose relative ranking was not maintained across all the datasets sizes we trained on. 
We therefore suspect that, in expectation, whenever 
a model A achieves better generalization performance than an architecture 
B given a certain sized training set, it will also have a better performance
for all other training set sizes.

\Cref{fig:imagenet-compare} (top-row) shows the corresponding results for ImageNet, with uncertainty estimates computed over 5 seeds. 
All models here are slightly oversized versions of their literature counterparts, for example
the ResNets use 384 channels in their lowest ResNet block instead of
256 like the ones described by \citet{he2016deep}.

Not only the standard image recognition models maintain their relative ranking as the 
training set size changes, but even the S3TA model follows the same pattern.
Indeed we notice that the generalization performance-curve is dominated by the ResNet-style feature
extractor, i.e, that training a S3TA model with a ResNet-50 style feature extractor moves the curve close to the standard ResNet-50 performance curve.
In the interest of clarity, to better appreciate the difference between the curves, we plot them in a separate figure without uncertainty estimates (see Supplementary material).

\Cref{fig:cifar10-compare} (top-row) shows the results for models on CIFAR10, with uncertainty 
bands obtained from training 30 different seeds. Even though the uncertainty is higher and the 
models are generally closer together, we still observe that the model 
architectures maintain their relative ranking, within their respective uncertainty bands.
For all these models the generalization error seems to be far away from converging to 
the underlying (irreducible) Bayes error even when training on the full dataset.

\subsection{Uncertainty of model selection}

If the ranking is preserved across dataset sizes we should be able to perform model 
selection using training sets of arbitrary size and assume that our decision
remains optimal for larger datasets created from the same underlying distribution.

Visually inspecting  \Cref{fig:mnist-compare}, \ref{fig:imagenet-compare}
or \ref{fig:cifar10-compare} reveals indeed some of the features we expect in this case: 
In the limit of very small datasets, all models start on average 
with a uniform prediction, achieving a generalization 
loss of $\approx \log(\text{\#classes})$ for balanced datasets.
In the limit of very large datasets, and assuming sufficiently 
large models and effective training procedures, they all approach a
common performance: presumably the irreducible Bayes error of the 
dataset. For intermediate sized training sets though, different models
exhibit vastly more varied generalization performance. 
In fact, we see that the performance gap between any two models is visibly 
larger for some intermediate sized training sets than for the full sized datasets.

While this is encouraging, optimization of neural networks is a stochastic process, with noise  
from the selection of training and calibration examples, from parameter 
initialization, the mini-batching process and potentially others sources. 
To ensure a reliable model selection process one can not consider the absolute performance gap alone,
but should also take into account its uncertainty.
To quantify the reliability of the model selection signal, i.e. the performance gap, 
we estimate the signal to noise ratio:
$SNR = \sqrt{\frac{\Delta^2}{Var[\Delta]}}$ with $\Delta = \LL_A - \LL_B$, 
where $\LL_X$ is the final generalization cross-entropy after training model $X$. 
The bottom rows of \Cref{fig:mnist-compare,fig:imagenet-compare} show the 
SNR estimates computed using 1000 bootstrap samples.

\subsection{Computational savings from smaller datasets}

We used fixed learning rates, or fixed learning rate schedules, for all
our experiments without any early stopping. 
This simplified the experimental setup and reduced the number of additional hyperparameters we needed to tune independently per dataset size, which could have skewed the results if not done fairly.
\jorg{Include this? Remove it?}\simon{Happy to keep it; but a good candidate for removal if we get tight on space?}
But that also means that, by choice, we did not enjoy any computational savings
from using smaller datasets, even though \Cref{fig:learningcurves} suggests 
that it takes fewer training steps to converge on smaller datasets and thus that such savings should be possible.

To get a lower bound on the potential computational savings, we implemented a
simple automatic annealing scheme for ImageNet: Every $1.28M / 256 = 5000$ 
training steps we performed a full-batch evaluation of the calibrated cross-entropy 
on the calibration set. We lowered the learning rate by a factor of 10 
when the performance did not improve for three consecutive iterations. We terminated training when the learning rate 
reached $\sfrac{1}{1000}$-th of it's initial value.
Using this schedule for ResNet models, we use 128k training examples 
(which provides a good SNR, as suggested by \Cref{fig:imagenet-compare}) 
and observe that training terminates after 4-7x fewer gradient steps 
than training on the full 1.28M examples.

%% file: 04-mdl.tex
\section{MDL and Bayes Factor Estimation}
\label{mdl}

We now show that we can use calibrated generalization cross-entropy estimates 
to perform model selection following {\em Occam's Razor}, 
the principle according to which we should prefer simpler models if possible.

Within the field of deep learning model selection is typically done with a simple
cross validation strategy: Split the available data into a training and a validation
set and choose the model that performs best on the validation subset. 
This approach works well when a lot of data is available and when overfitting 
is not a major concern.

Other approaches to model selection like for example Bayesian model selection or the
Minimum Description Length (MDL) principle have been studied in detail and can often be 
understood in a context of a general theory of inductive inference. For these it is known
that their definition gives raise to a notion of the models {\em complexity}. And we know 
that they prefer to select simpler models (according to their respective complexity measure) 
if they have similar predictive performance.

Unfortunately, neither of these approaches are naively applicable to 
deep neural networks: To perform Bayesian model selection between 
two model architectures $\HH_1$ and $\HH_2$ for example, we need to compute
$\frac{p(\HH_1 | \DD)}{p(\HH_2 | \DD)} = \frac{p(\DD | \HH_1) p(\HH_1)}{p(\DD | \HH_2) p(\HH_2)}$,
where $p(\DD | \HH_i) = \int_\Theta p(\DD | \Theta) p(\Theta) d\Theta$, which is in most cases intractable.
Indeed, estimating this quantity for deep neural networks proved to 
to be a very challenging problem, and usually requires changes to the 
parameterization and training procedure, often leading to 
models with significantly worse predictive performance \citep{mackay2003information}.

Minimum Description Length (MDL) \citep{rissanen1978modeling,rissanen1989stochastic,grunwald2007mdl} provides
a closely related alternative principle that also
considers $p(\DD|\HH_i)$ the quantity of interest, but does not force us to consider the prior 
$p(\Theta)$ and posterior $p(\Theta | \DD)$ explicitly. 
Instead, we are only concerned with the shortest possible compression of 
the dataset $\DD$: 
The length of a message that transmits the content of the dataset (the labels) uniquely 
from a sender to a receiver assuming some model and coding-procedure $\HH$.
We will refer to the length of the message as $\MDL(\DD | \HH)$.
By virtue of the Kraft-McMillan theorem, we know that any such  
procedure corresponds to a probability distribution over all possible 
datasets: since $\sum_\DD \exp(-\MDL(\DD | \HH)) \leq 1$, we can identify
$p(\DD | \HH) \propto \exp(-\MDL(\DD | \HH))$.

\begin{table}
    {\scriptsize
    \begin{tabular}{ l | c | c | c | c}
    $\MDL(\DD | \HH_1) - \MDL(\DD | \HH_2)$ & LR & MLP & MLP+do  \\ 
    \hline
    Log. regression &  &  & \\
    MLP           & $11982 \pm 142$ &  &  \\   %
    MLP+dropout   & $12237 \pm 154$ &  $254 \pm 60$ &  \\  %
    ConvNet       & $15293 \pm 126$ &  $3310 \pm 64$ &  $3057 \pm 87$ \\   %
    \end{tabular}
    }
    \caption{
    MDL log-evidence estimates for various model pairs on MNIST in nats. 
    For example, the evidence in favour of an MLP with dropout vs an MLP without is $\exp(254) \approx 10^{110} $ to 1.
    }
    \label{mdl:emnist}
\end{table}

\citet{blier2018description} described an approach to constructing a
{\em prequential} \citep{dawid1984present, grunwald2007mdl} code that 
works well together with deep neural networks:
The basic idea is to transmit one datapoint at a time; always coding
it with a code derived from a model that was trained on the previously 
transmitted datapoints. When the sender and receiver use a deterministic 
training method (i.e. fixed seeds for all randomness; well-defined stopping 
criteria etc.) both sides learn exactly the same model and therefore
can construct a common code from its predictions. The total length of the message 
to transmit a dataset $\DD = \{y_i\}_{i=1}^N$ of size $N$ is therefore
$\MDL(\DD|\HH) \leq \sum_{i=1}^N \log p(y_i | \{y_j\}_{j<i}, \HH)$. 
\citet{blier2018description} showed that they obtain significantly 
shorter description lengths than with other approaches to estimate 
them for deep learning models. 
However, they did not have a principled way to regularize the models to be 
trainable on small datasets. 
Instead they proposed to switch between different model architectures 
with different capacities; therefore not obtaining description lengths 
for a single model, but rather for the entire procedure consisting of multiple, 
increasingly more elaborate architectures and their switching pattern. 

Using calibration we can instead construct a
coding method for a single, fixed model and training procedure: 
Both sender and receiver simply derive a training and calibration
split from the previously transmitted data and use the calibrated 
cross-entropy prediction to construct the code for the next datapoint.
The remainder of the prequential approach remains unchanged.

In practice this means that the description length of a dataset for a given 
$\HH$ can be approximated by the {\em area under the curve} of 
plots like \Cref{fig:mnist-compare}, \ref{fig:imagenet-compare} and \ref{fig:cifar10-compare},
if we just construct the \emph{available}  
and \emph{evaluation} sets in a specific way: Instead of randomly sampling them 
from the underlying distribution, we have to ensure that the available 
dataset for a subset of size $N > M$ is a superset of the datapoints used 
when training the model for the subset of size $M$; and the evaluation examples 
have to be exactly those examples that were added.

A description length based model selection approach thus considers the area 
between two plots in \Cref{fig:mnist-compare,fig:imagenet-compare}
and \Cref{fig:cifar10-compare} (top-row) because it corresponds to the 
log-evidence in favour of the model with the smaller area under the curve:
$\log \MDL(\DD|\HH_1) - \log \MDL(\DD|\HH_2)$.

\begin{table}
    {\scriptsize
    \begin{tabular}{ l | c | c | c | c}
    $\MDL(\DD | \HH_1) - \MDL(\DD | \HH_2)$ & LR & MLP & MLP+do  \\ 
    \hline
    MLP           & $11982 \pm 142$ &  &  \\
    MLP+dropout   & $12237 \pm 154$ &  $254 \pm 60$ &  \\
    ConvNet       & $15293 \pm 126$ &  $3310 \pm 64$ &  $3057 \pm 87$ \\ 
    \end{tabular}
    \vspace{-0.2cm}
    }
    \caption{
    MDL log-evidence estimates for selected model pairs on CIFAR10 in nats.}
    \label{mdl:cifar10} 
    \vspace{-0.2cm}    
\end{table}

\begin{table}
    {\footnotesize
    \begin{tabular}{ l | c | c | c | c}
    {\scriptsize $\MDL(\DD | \HH_1) - \MDL(\DD | \HH_2)$} & VGG16 & RN-50 & RN-101  \\ 
    \hline
    ResNet-50     & $1.14 M$ &  &  \\
    ResNet-101    & $1.23 M$ &  $87 k$ &  \\
    ResNet-152    & $1.27 M$ &  $125 k$ &  $37 k$ \\ 
    \end{tabular}
    \vspace{-0.1cm}
    }
    \caption{MDL log-evidence estimates for selected model pairs on ImageNet in nats. 
        The uncertainty for all models is less then 20k nats for all pairings.}
    \label{mdl:imagenet}
    \vspace{-0.1cm}    
\end{table}

We implement the suggested splitting procedure and estimate the area 
between the performance curves using a simple trapezoid integral 
approximation for various model pairs. We obtain uncertainty estimates on this measure
by running 3 different seeds and by varying the number points for the 
interval estimate (\Cref{mdl:emnist,mdl:cifar10,mdl:imagenet}).

%% file: 05-conclusions.tex
\section{Conclusions}

In this paper we present a comprehensive empirical study 
on how overparameterized neural networks generalize as a function of
the training set size. Our results confirm and extend the recently 
growing body of literature that analyzes and seeks to understand the curiously 
good generalization performance of big neural networks; and takes an in-depth analysis of 
the dependency between the training set size and the generalization performance.

From our experimental result we derive the hypothesis that sufficiently big 
neural networks, those that operate far beyond their interpolation threshold, 
maintain their expected relative generalization performance ranking irrespective of the size of the training set they are trained on. While we are not aware of a theoretical argument that could support this conjecture, we provide a strong empirical verification of this claim conducted on several architectures, model sizes, datasets and optimization methods.
The results of our experiments have a wide range of practical implications for model selection and architecture search, even more so because our experiments show that training on smaller subsets of the data can not only save computational resources, but also improve the model selection signal at the same
time. Both these aspects are of particular importance for neural architecture search approaches
and can find immediate practical application. 

Finally, in this work we also show that simple temperature
calibration is sufficient to obtain reliable and well behaved generalization 
performance in terms of cross-entropy. Taking a prequential perspective, this allows us to 
compute Minimum Description Length estimates for deep-learning models, 
a quantity that was previously inaccessible, and thereby paving the way for a principled 
model selection procedure that takes into account Occam's-razor.

\section*{Acknowledgements}

We would like to thank Andriy Mnih, Markus Wulfmeier, Sam Smith and Yee Whye Teh for helpful discussions and their insightful comments while working on this paper.

%% file: supplement.tex
\newpage
\onecolumn
\section*{Supplement}

\subsection*{Scaling the Model Size}

Cross-entropy performance profiles for various models and 
datasets. We always use 90\% of the available data for training, 10\% 
for calibration and evaluate on the (official) validation/development 
sets provided by the different datasets.

\subsubsection*{Fully Connected MLP on MNIST}
\includegraphics[width=\linewidth]{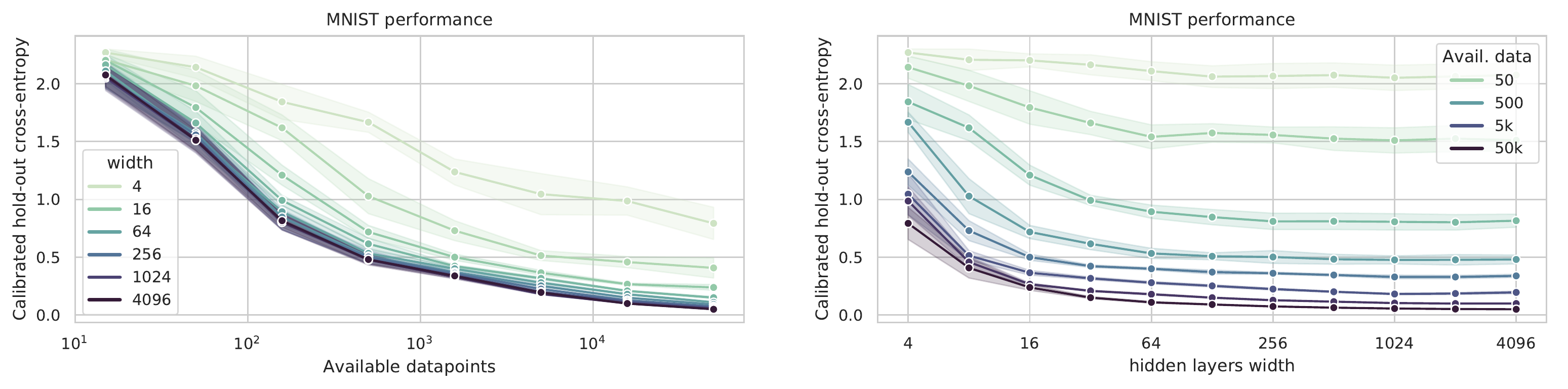}

\subsubsection*{Fully Connected MLP on EMNIST} 
\includegraphics[width=\linewidth]{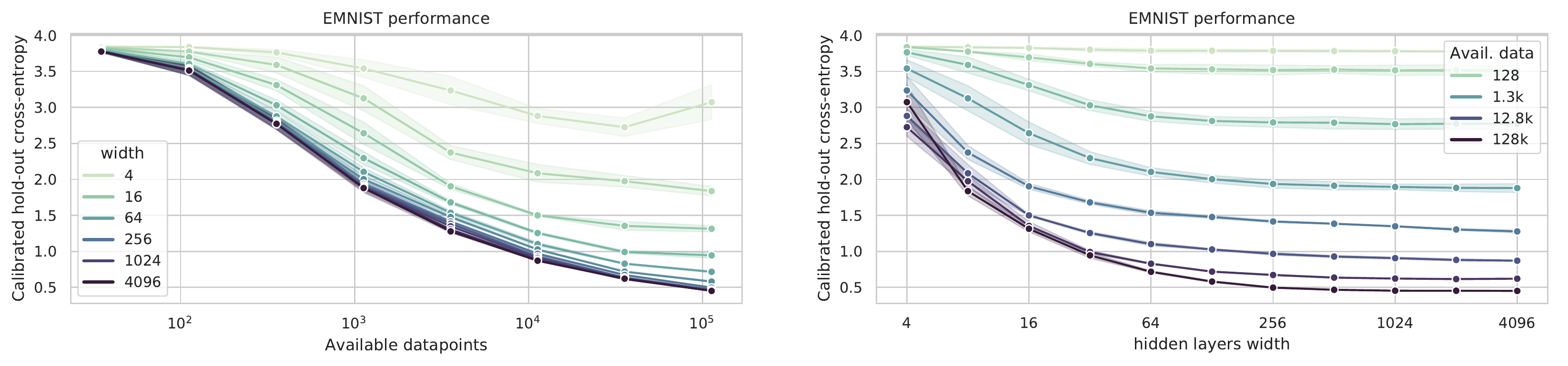}

\subsubsection*{Fully Connected MLP on CIFAR10}
\includegraphics[width=\linewidth]{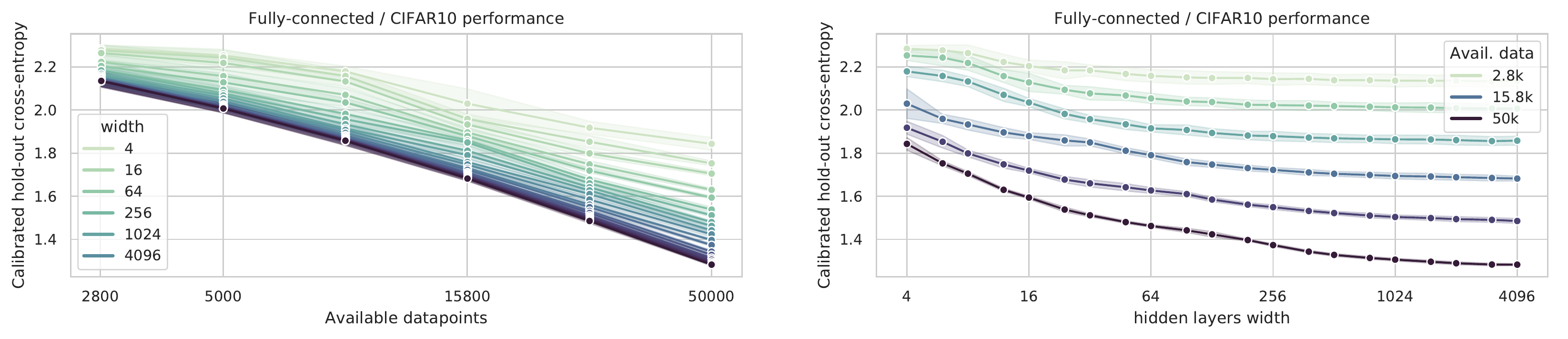}

\subsubsection*{ResNet-20 on CIFAR10}
\includegraphics[width=\linewidth]{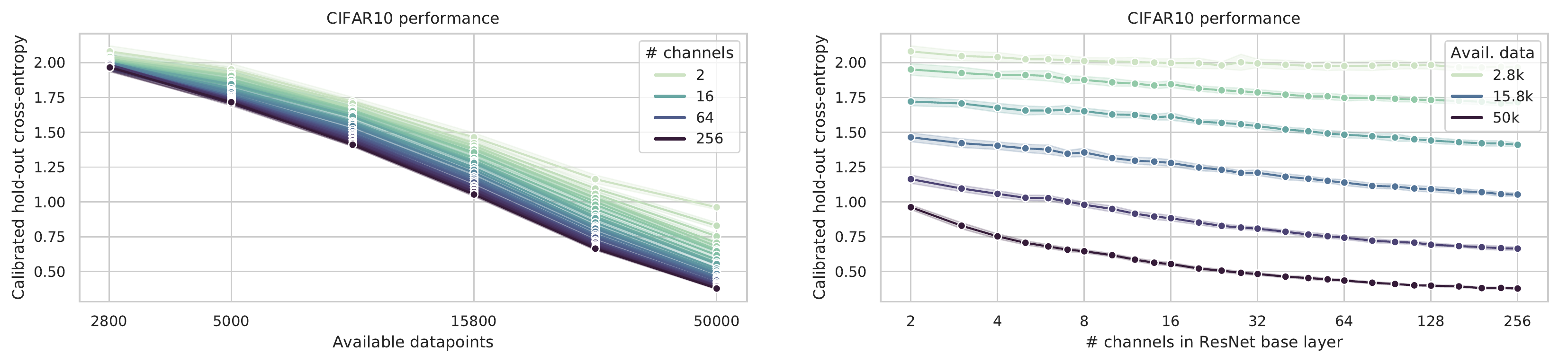}

\subsubsection*{ResNet-101 on ImageNet}
\includegraphics[width=\linewidth]{fig-channels-imagenet}

\subsection*{Calibrated Cross-Entropy vs. Error-rates}

\subsubsection*{EMNIST model comparison with either cross-entropy or error-rate}
\includegraphics[width=\linewidth]{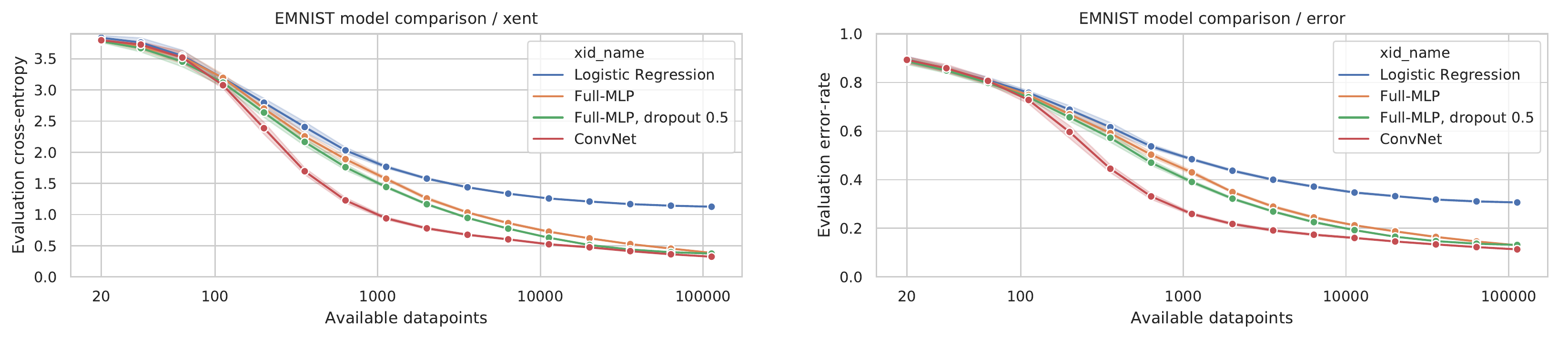}

\subsubsection*{ImageNet model comparison with either cross-entropy or error-rate}
\includegraphics[width=\linewidth]{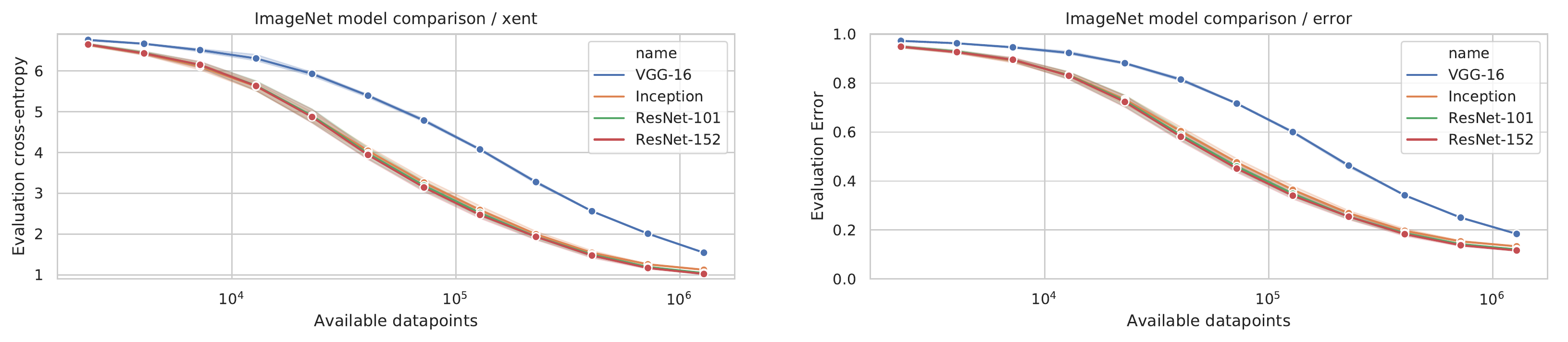}

\subsection*{S3TA on ImageNet}

We here compare the sequential, attention based {\em S3TA} model
against various standard architectures for ImageNet. 
We use 8 sequential attention steps and a Resnet101-based feature extractor.

\vspace{1em}
\includegraphics[width=\linewidth]{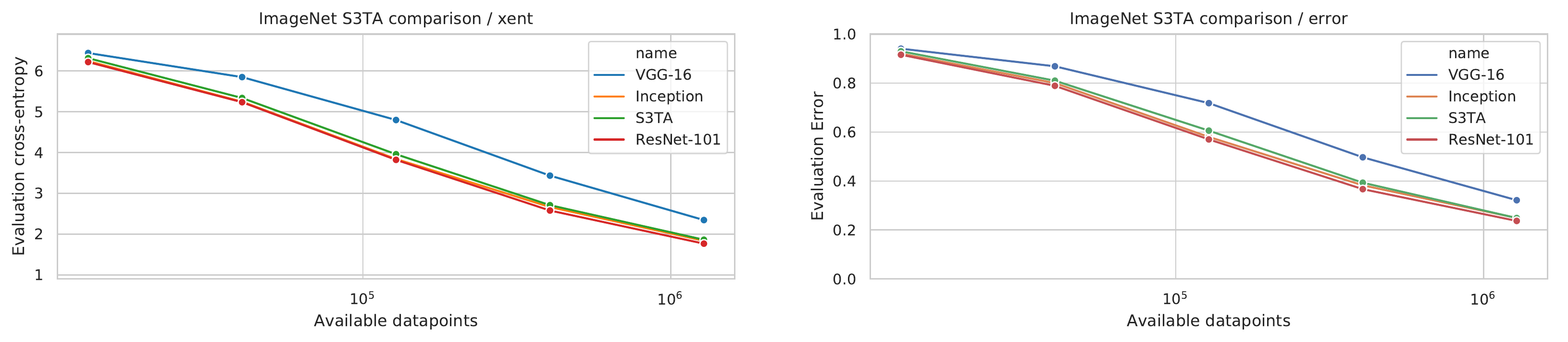}

\subsection*{Compute Infrastructure and Experimental details}

We implemented all experiments in Tensorflow and use existing, publicly available code wherever possible. 
E.g. we use the existing open-source implementation of the 
NASBench-101 architectures and of Inception; 
custom implementations of MLPs, ConvNets and ResNets.
MLPs where executed on CPUs, simple ConvNets on single GPUs and
bigger ResNet, Inception and S3TA models on 4 or 8 TPUs synchronously in parallel.
The total batch-size was always fixed to 256.

\subsection*{NASBench architectures}

These are the NAS-Bench 101 architectures considered in the paper, with their corresponding hashes. We picked architectures equidistantly in terms of performance from the BASBench database, after disregarding
the worst 10\%.

\subsubsection*{75ddc0891320c863ec5f148ae675947e}
['input', 'conv1x1-bn-relu', 'conv3x3-bn-relu', 'conv1x1-bn-relu', 'maxpool3x3', 'maxpool3x3', 'output']
\begin{align}
M = 
\begin{bmatrix}
0 & 1 & 0 & 0 & 0 & 0 & 0 \\
0 & 0 & 1 & 1 & 0 & 0 & 0 \\
0 & 0 & 0 & 1 & 1 & 0 & 0 \\
0 & 0 & 0 & 0 & 0 & 1 & 1 \\
0 & 0 & 0 & 0 & 0 & 1 & 0 \\
0 & 0 & 0 & 0 & 0 & 0 & 1 \\
0 & 0 & 0 & 0 & 0 & 0 & 0 \\
\end{bmatrix}
\end{align}

\subsubsection*{b5a2bfe35a8f6a21364a992d4dadad31}
['input', 'maxpool3x3', 'maxpool3x3', 'conv3x3-bn-relu', 'maxpool3x3', 'conv3x3-bn-relu', 'output']
\begin{align}
M = 
\begin{bmatrix}
0 & 1 & 0 & 0 & 0 & 0 & 1 \\
0 & 0 & 1 & 0 & 0 & 0 & 0 \\
0 & 0 & 0 & 1 & 1 & 0 & 1 \\
0 & 0 & 0 & 0 & 0 & 0 & 1 \\
0 & 0 & 0 & 0 & 0 & 1 & 0 \\
0 & 0 & 0 & 0 & 0 & 0 & 1 \\
0 & 0 & 0 & 0 & 0 & 0 & 0 \\
\end{bmatrix}
\end{align}

\subsubsection*{63e9304e6a2aa542eb273dce26477c38}
['input', 'maxpool3x3', 'conv3x3-bn-relu', 'conv3x3-bn-relu', 'maxpool3x3', 'maxpool3x3', 'output']
\begin{align}
M = 
\begin{bmatrix}
0 & 1 & 0 & 0 & 0 & 1 & 1 \\
0 & 0 & 1 & 1 & 0 & 0 & 0 \\
0 & 0 & 0 & 0 & 0 & 0 & 1 \\
0 & 0 & 0 & 0 & 1 & 0 & 0 \\
0 & 0 & 0 & 0 & 0 & 1 & 0 \\
0 & 0 & 0 & 0 & 0 & 0 & 1 \\
0 & 0 & 0 & 0 & 0 & 0 & 0 \\
\end{bmatrix}
\end{align}

\subsubsection*{9f5da3119e80518bd23f9c115c7a18d6}
['input', 'conv3x3-bn-relu', 'conv1x1-bn-relu', 'conv1x1-bn-relu', 'conv1x1-bn-relu', 'conv3x3-bn-relu', 'output']
\begin{align}
M = 
\begin{bmatrix}
0 & 1 & 0 & 1 & 0 & 0 & 0 \\
0 & 0 & 1 & 0 & 0 & 1 & 0 \\
0 & 0 & 0 & 1 & 0 & 0 & 1 \\
0 & 0 & 0 & 0 & 1 & 0 & 0 \\
0 & 0 & 0 & 0 & 0 & 1 & 0 \\
0 & 0 & 0 & 0 & 0 & 0 & 1 \\
0 & 0 & 0 & 0 & 0 & 0 & 0 \\
\end{bmatrix}
\end{align}

\subsubsection*{0da48e9f9faecf504244c65b82e0ba71}
['input', 'conv3x3-bn-relu', 'conv3x3-bn-relu', 'conv3x3-bn-relu', 'conv3x3-bn-relu', 'output']
\begin{align}
M = 
\begin{bmatrix}
0 & 1 & 1 & 1 & 1 & 1 \\
0 & 0 & 0 & 0 & 1 & 0 \\
0 & 0 & 0 & 1 & 0 & 0 \\
0 & 0 & 0 & 0 & 1 & 0 \\
0 & 0 & 0 & 0 & 0 & 1 \\
0 & 0 & 0 & 0 & 0 & 0 \\
\end{bmatrix}
\end{align}